\documentclass{article}

\usepackage{amsmath,amsfonts,bm}

\def\eqref#1{equation~\ref{#1}}

\def\1{\bm{1}}

\DeclareMathAlphabet{\mathsfit}{\encodingdefault}{\sfdefault}{m}{sl}
\SetMathAlphabet{\mathsfit}{bold}{\encodingdefault}{\sfdefault}{bx}{n}

\usepackage{microtype}
\usepackage{graphicx}
\usepackage{subfigure}
\usepackage{booktabs} 
\usepackage{mathtools}
\usepackage{tikz}
\usepackage{amssymb}
\usepackage{amsthm}
\usepackage{anyfontsize}
\usepackage{multirow}

\usepackage{hyperref}

\newcommand{\minihead}[1]{{\vspace{.45em}\noindent\textbf{#1.}\hspace{0.5em}}}

\newcommand{\img}{\phi}
\newcommand{\Img}{\Phi}
\newcommand{\pose}{\theta}
\newcommand{\posefn}{f}

\newcommand{\et}{L}
\newcommand{\cc}{\rho}

\newtheorem{proposition}{Proposition}

\newcommand{\Arrow}[1]{%
\parbox{#1}{\tikz{\draw[->](0,0)--(#1,0);}}
}

\usepackage[accepted]{icml2019}

\icmltitlerunning{Equivariant Transformer Networks}

\begin{document}

\twocolumn[
\icmltitle{Equivariant Transformer Networks}



\icmlsetsymbol{equal}{*}

\begin{icmlauthorlist}
\icmlauthor{Kai Sheng Tai}{stanford}
\icmlauthor{Peter Bailis}{stanford}
\icmlauthor{Gregory Valiant}{stanford}
\end{icmlauthorlist}

\icmlaffiliation{stanford}{Stanford University, Stanford, CA, USA}

\icmlcorrespondingauthor{Kai Sheng Tai}{kst@cs.stanford.edu}


\vskip 0.3in
]



\printAffiliationsAndNotice{}  

\begin{abstract}
How can prior knowledge on the transformation invariances of a domain be incorporated into the architecture of a neural network?
We propose Equivariant Transformers (ETs), a family of differentiable image-to-image mappings that improve the robustness of models towards pre-defined continuous transformation groups.
Through the use of specially-derived canonical coordinate systems, ETs incorporate functions that are {equivariant} \emph{by construction} with respect to these transformations.
We show empirically that ETs can be flexibly composed to improve model robustness towards more complicated transformation groups in several parameters.
On a real-world image classification task, ETs improve the sample efficiency of ResNet classifiers, achieving relative improvements
in error rate of up to $15\%$ in the limited data regime while increasing model parameter count by less than $1\%$.

\end{abstract}

\section{Introduction}
\label{sec:introduction}

In computer vision, we are often equipped with prior knowledge on the transformation invariances of a domain.
Consider, for example, the problem of classifying street signs in real-world images.
In this domain, we know that the appearance of a sign in an image is subject to various deformations: the sign may be rotated, its scale will depend on its distance, and it may appear distorted due to perspective in 3D space.
Regardless, the identity of the street sign should remain invariant to these transformations.

With the exception of translation invariance, convolutional neural network (CNN) architectures typically do not take advantage of such prior knowledge on the transformation invariances of the domain.
Instead, current standard practice heuristically incorporates these priors during training via data augmentation (\emph{e.g.}, by applying a random rotation or scaling to each training image).
While data augmentation typically helps reduce the test error of CNN-based models, there is no guarantee that transformation invariance will be enforced for data not seen during training.

In contrast to training time approaches like data augmentation, recent work on group equivariant CNNs \citep{cohen2016group,dieleman2016exploiting,marcos2017rotation,worrall2017harmonic,henriques2017warped,cohen2018spherical} has explored new CNN architectures that are guaranteed to respond predictably to particular transformations of the input.
For example, the CNN model family may be constrained such that a rotation of the input results in a corresponding rotation of its subsequent representation, a property known as equivariance.
However, these techniques---most commonly designed for rotations and translations of the input (\emph{e.g.}, \citet{dieleman2016exploiting,marcos2017rotation,worrall2017harmonic})---fail to generalize to deeper compositions of continuous transformations.
This limits the applicability of these techniques in more complicated real-world scenarios involving continuous transformations in several dimensions, such as the above example of street sign classification.

To address these shortcomings of group equivariant CNNs, we propose \emph{Equivariant Transformer} (ET) layers, a flexible class of functions that improves robustness towards arbitrary pre-defined groups of continuous transformations.
An ET layer for a transformation group $G$ is an image-to-image mapping that satisfies the following local invariance property: 
for any input image $\phi$ and transformation $T\in G$, the images $\phi$ and $T\phi$ are both mapped to the same output image.
ET layers are differentiable with respect to both their parameters and input, and thus can be easily incorporated into existing CNN architectures.
Additionally, ET layers can be flexibly combined to achieve improved invariance towards more complicated compositions of transformations (\emph{e.g.}, simultaneous rotation, scale, shear, and perspective transformations).

Importantly, the invariance property of ETs holds \emph{by construction}, without any dependence on additional heuristics during training.
We achieve this by using the method of \emph{canonical coordinates} for Lie groups~\citep{rubinstein1991recognition}.
The key property of canonical coordinates that we utilize is their ability to reduce arbitrary continuous transformations to translation.
For example, polar coordinates are canonical coordinates for the rotation group, since a rotation reduces to a translation in the angular coordinate.
These specialized coordinates can be analytically derived for a given transformation and efficiently implemented within a neural network.

We evaluate the performance of ETs using both synthetic and real-world image classification tasks.
Empirically, ET layers improve the sample efficiency of image classifiers relative to standard Spatial Transformer layers~\citep{jaderberg2015spatial}.
In particular, we demonstrate that ET layers improve the sample efficiency of modern ResNet classifiers on the Street View House Numbers dataset, with relative improvements in error rate of up to $15\%$ in the limited data regime.
Moreover, we show that a ResNet-10 classifier augmented with ET layers is able to exceed the accuracy achieved by a more complicated ResNet-34 classifier without ETs,  thus reducing both memory usage and computational cost.

\section{Related Work}
\label{sec:related}

\vspace{-3pt}
\minihead{Equivariant CNNs} There has been substantial recent interest in CNN architectures that are equivariant with respect to transformation groups other than  translation.
Equivariance with respect to discrete transformation groups (\emph{e.g.}, reflections and $90^\mathrm{o}$ rotations) can be achieved by transforming CNN filters or feature maps using the group action \citep{cohen2016group,dieleman2016exploiting,laptev2016ti,marcos2017rotation,zhou2017oriented}.
Invariance can then be achieved by pooling over this additional dimension in the output of each layer.
In practice, this technique supports only relatively small discrete groups since its computational cost scales linearly with the cardinality of the group.

Methods for achieving equivariance with respect to continuous transformation groups fall into one of two classes: those that expand the input in a \emph{steerable basis}~\citep{amari1978feature,freeman1991design,teo1998theory,worrall2017harmonic,jacobsen2017dynamic,weiler2018learning,cohen2018spherical}, and those that compute convolutions under a specialized \emph{coordinate system}~\citep{rubinstein1991recognition,segman1992canonical,henriques2017warped,esteves2018polar}.
The relationship between these two categories of methods is analogous to the duality between frequency domain and time domain methods of signal analysis.
Our work falls under the latter category that uses coordinate systems specialized to the transformation groups of interest.

\vspace{-3pt}
\minihead{Equivariance via Canonical Coordinates} 
\citet{henriques2017warped} apply CNNs to images represented using coordinate grids computed using a given pair of continuous, commutative transformations.
Closely related to this technique are Polar Transformer Networks~\citep{esteves2018polar}, a method that handles images deformed by translation, rotation, and dilation by first predicting an origin for each image before applying a CNN over log-polar coordinates.
Unlike these methods, we handle higher-dimensional transformation groups by passing an input image through a sequence of ET layers in series.
In contrast to \citet{henriques2017warped}, where a pair of commutative transformations is assumed to be given as input, we show how canonical coordinate systems can be analytically derived given only a single one-parameter transformation group using technical tools described by \citet{rubinstein1991recognition}.


\vspace{-3pt}
\minihead{Spatial Transformer Networks}
As with Spatial Transformer (ST) layers~\citep{jaderberg2015spatial}, our ET layers aim to factor out nuisance modes of variation in images due to various geometric transformations.
Unlike STs, ETs incorporate additional structure in the functions used to predict transformations.
We expand on the relationship between ETs and STs in the following sections.

\vspace{-3pt}
\minihead{Locally-Linear Approximations} 
\citet{gens2014deep} use local search to approximately align filters to image patches, in contrast to our use of a global change of coordinates.
The sequential pose prediction process in a stack of ET layers is also reminiscent of  the iterative nature of the Lucas-Kanade (LK) algorithm and its descendants~\citep{lucas1981iterative,lin2017inverse}. 

\vspace{-3pt}
\minihead{Image Registration and Canonicalization}
ETs are related to classic ``phase correlation'' techniques for image registration that compare the Fourier or Fourier-Mellin transforms of an image pair~\citep{de1987registration,reddy1996fft}; these methods can be interpreted as Fourier basis expansions under canonical coordinate systems for the relevant transformations.
Additionally, the notion of image canonicalization relates to work on \emph{deformable templates}, where object instances are generated via deformations of a prototypical object \citep{amit1991structural,yuille1991deformable,shu2018deforming}.

\section{Problem Statement}
\label{sec:problem}

In this section, we begin by reviewing influential prior work on image canonicalization with Spatial Transformers \citep{jaderberg2015spatial}. 
We then argue that the lack of \emph{self-consistency} in pose prediction is a key weakness with the standard ST that results in poor sample efficiency.

\subsection{Image Canonicalization with Spatial Transformers}
\label{sec:problem-canonicalization}

Suppose that we observed a collection of images $\phi(\mathbf{x})$, each of which is a mapping from image coordinates $\mathbf{x}\in\mathbb{R}^2$ to pixel intensities in each channel.
Each image is a transformed version of some latent \emph{canonical} image $\img_*$: $\img = T_\theta \img_* \coloneqq \img_*(T_\theta \mathbf{x})$, where the transformation $T_\theta:\mathbb{R}^2\rightarrow\mathbb{R}^2$ is modulated by  \emph{pose parameters} $\theta\in\mathbb{R}^k$.

\begin{figure*}[t]
\centering
\includegraphics[width=0.85\textwidth]{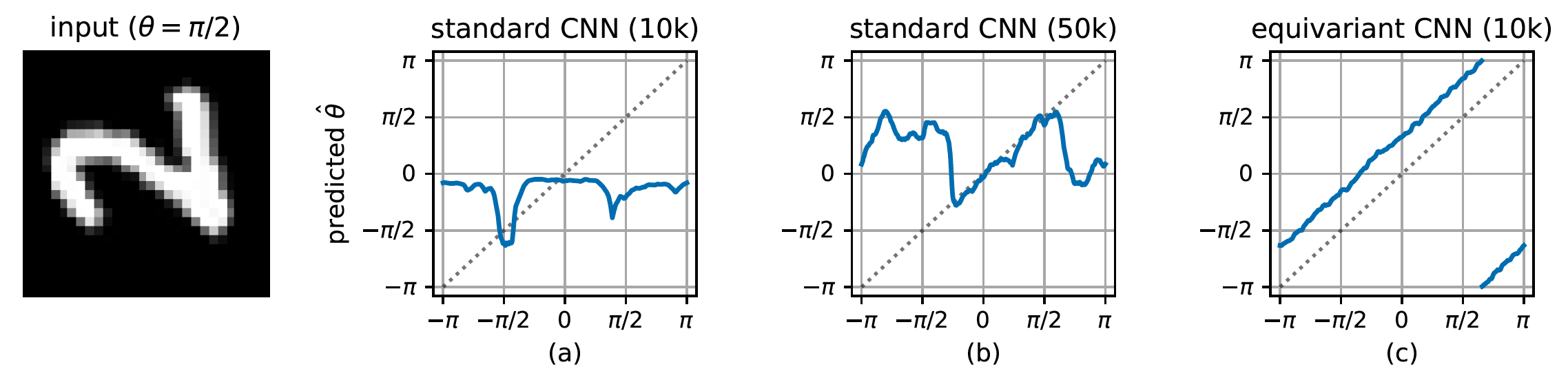}
\vspace{-1em}
\caption{
\textbf{Sample complexity for predicting rotations.}
Predicted rotation angles vs. true angles for a rotated MNIST digit (left).
The predictions of a self-consistent pose predictor will be parallel to the diagonal (dotted line).
\textbf{(a)} After training with 10k rotated examples, a pose prediction CNN is not self-consistent; \textbf{(b)} with 50k rotated examples, it is only self-consistent over a limited range of angles. 
In contrast, \textbf{(c)} a rotationally-equivariant CNN outputs self-consistent predictions after 10k examples (with small error due to interpolation and boundary effects).
There is a nonzero bias in $\hat\pose$ since the pose labels are latent and there is no preferred image orientation.
}
\label{fig:equivariant-vs-nonequivariant}
\vspace{-8pt}
\end{figure*}

If the transformation family and the pose parameters $\pose$ for each image $\img$ are known, then the learning problem may be greatly simplified.
If $T_\pose$ is invertible, then access to $\pose$ implies that we can recover $\img_*$ from $\img$ via $T^{-1}_\pose \img = T_\pose^{-1} T_\pose \img_* = \img_*$. 
This is advantageous for learning when $\img_*$ is drawn from a small or even finite set (\emph{e.g.}, $\phi_*$ could be sampled from a finite set of digits, while $\phi$ belongs to an infinite set of transformed images).

When the pose parameters are latent, as is typical in practice, we can attempt to predict an appropriate inverse transformation from the observed input.\footnote{For example, the apparent convergence of parallel lines in the background of an image can provide information on the correct inverse projective transformation to be applied.}
Based on this intuition, a Spatial Transformer (ST) layer $L: \Img \rightarrow \Img$ \citep{jaderberg2015spatial} transforms an input image $\img$ using pose parameters $\hat\pose = \posefn(\phi)$ that are predicted as a function of the input:
\begin{equation*}
  L(\phi) = T^{-1}_{f(\phi)} \phi,
\end{equation*} 
where the \emph{pose predictor} $\posefn : \Img \rightarrow \mathbb{R}^k$ is typically parameterized as a CNN or fully-connected network.




%

\vspace{-2pt}
\subsection{Self-Consistent Pose Prediction}
\label{sec:problem-consistency}

A key weakness of standard STs is the pose predictor's lack of robustness to transformations of its input.
As a motivating example, consider images in a domain that is known to be rotationally invariant (\emph{e.g.}, classification of astronomical objects), and  
suppose that we train an ST-augmented CNN that aims to canonicalize the rotation angle of input images. 
For some input $\img$, let the output of the pose predictor be $\posefn(\img) = \hat\pose$ for some $\hat\pose \in [0, 2\pi)$.
Then given $T_{\pose} \phi$ (\emph{i.e.}, the same image rotated by an additional angle $\theta$), we should expect the output of an ideal pose predictor to be $\posefn(T_{\pose} \img) = \hat\pose + \pose + 2\pi m$ for some integer $m$.
In other words, the pose prediction for an input $\phi$ should constrain those for $T_\pose \phi$ over the entire orbit of the transformation.

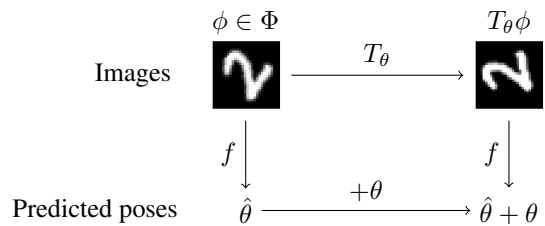
\begin{figure}
\centering
\begin{tikzpicture}
\node[] (a) at (0, 0) {\includegraphics[width=2.75em]{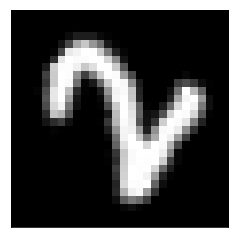}};
\node[] (b) at (3.5, 0) {\includegraphics[width=2.75em]{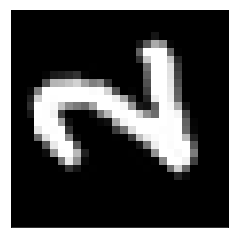}};
\node[] (a-cap) at (0, 0.7) {$\img\in\Img$};
\node[] (b-cap) at (3.5, 0.7) {$T_\theta \img$};
\node[] (c) at (0, -1.8) {$\hat\theta$};
\node[] (d) at (3.5, -1.8) {$\hat\theta + \theta$};
\node[] (images) at (-1.5, 0) {Images};
\node[] (poses) at (-2, -1.8) {Predicted poses};
\draw [->] (a) to node[above]{$T_\theta$} (b);
\draw [->] (a) to node[left]{$f$} (c);
\draw [->] (c) to node[above]{$+\theta$} (d);
\draw [->] (b) to node[left]{$f$} (d);
\end{tikzpicture}
\caption{
\textbf{Self-consistent pose prediction.}
We call a function $f:\Img\rightarrow\mathbb{R}^k$ \emph{self-consistent} if the action of a transformation $T_\pose$ on its input results in a corresponding increment of $\pose$ in its output.
Self-consistency is desirable for functions that predict the pose (\emph{e.g.}, rotation angle) of an object in an image.
}
\label{fig:self-consistency}
\vspace{-7pt}
\end{figure}

We refer to this desired property of the pose prediction function as \emph{self-consistency} (Figure~\ref{fig:self-consistency}).
In general, we say that a pose prediction function $f: \Phi \rightarrow \mathbb{R}^k$ is \emph{self-consistent} with respect to a transformation group $G$ parameterized by $\mathbf{\theta}\in\mathbb{R}^k$ if $f(T_\mathbf{\theta} \phi) = f(\phi) + \mathbf{\theta}$, for any image $\phi$ and transformation $T_\mathbf{\theta} \in G$.
We note that self-consistency is a special case of group equivariance.\footnote{A function $f$ is \emph{equivariant} with respect to the group $G$ if there exist transformations $T_g$ and $T'_g$ such that $f(T_g\img) = T'_g f(\img)$ for all $g \in G$ and $\phi \in \Phi$.}

However, there is no guarantee that self-consistency should hold when pose prediction is performed using a standard CNN or fully-connected network: while standard CNNs are equivariant with respect to translation, they are not equivariant with respect to other transformation groups~\citep{cohen2016group}.
In Figure~\ref{fig:equivariant-vs-nonequivariant}, we illustrate a simple example of this limitation of standard CNNs.
Using MNIST digits rotated by angles uniformly sampled in $\pose\in[0,2\pi)$, we train a CNN classifier with a ST layer that predicts the rotation angle of the input image.
During training, the model receives a rotated image as input along with the class label $y\in \{0, \dots, 9\}$; the true rotation angle $\pose$ is unobserved.
In this example task, we find that the poses predicted by the CNN are only approximately self-consistent within a small range of angles, even when the network is trained with 50,000 examples. 
In contrast, a rotation-equivariant CNN can achieve approximate self-consistency given only 10,000 training examples.




\section{Equivariant Transformers}
\label{sec:method}




Due to this weakness of standard CNN pose predictors, we will instead use functions that are guaranteed \emph{by construction} to satisfy self-consistency.
We achieve this by leveraging the translation equivariance of standard CNN architectures in combination with specialized \emph{canonical coordinate systems} designed for the particular transformation groups of interest.
Canonical coordinates allow us to reduce the problem of self-consistent prediction with respect to an \emph{arbitrary} continuous transformation group to that of self-consistent prediction with respect to the translation group.

We begin with preliminaries on canonical coordinates systems~(\S\ref{sec:method-coords}).
We then describe our proposed Equivariant Transformer architecture~(\S\ref{sec:method-et}).
Next, we describe how canonical coordinates can be derived for a given transformation~(\S\ref{sec:method-deriving}).
Finally, we describe how ET layers can be applied sequentially to handle compositions of several transformations~(\S\ref{sec:method-composition}) and cover implementation details~(\S\ref{sec:method-implementation}).

\subsection{Canonical Coordinate Systems for Lie Groups}
\label{sec:method-coords}

The method of canonical coordinates was first described by \citet{rubinstein1991recognition} and later developed in more generality by \citet{segman1992canonical} for the purpose of computing image descriptors that are invariant under the action of continuous transformation groups.


A \emph{Lie group} with parameters $\theta\in\mathbb{R}^k$ is a group of transformations of the form $T_\theta: \mathbb{R}^d \rightarrow \mathbb{R}^d$ that are differentiable with respect to $\theta$.
We let the parameter $\theta=0$ correspond to the identity element, $T_0 \mathbf{x} = \mathbf{x}$.
A canonical coordinate system for $G$ is defined by an injective map $\cc$ from Cartesian coordinates to the new coordinate system that satisfies
\begin{equation}
  \cc(T_\theta\mathbf{x}) = \cc(\mathbf{x}) + \sum_{i=1}^k \theta_i \mathbf{e}_k,
  \label{eq:cc-condition}
\end{equation}
for all $T_\theta \in G$, where $\mathbf{e}_i$ denotes the $i$th standard basis vector.
Thus, a transformation by $T_\theta$ appears as a translation by $\theta$ under the canonical coordinate system.
To help build intuition, we give two examples of canonical coordinates:

\minihead{Example 1 (Rotation)} For $T_\theta \mathbf{x} = (x_1 \cos\theta - x_2\sin\theta, x_1\sin\theta + x_2\cos\theta)$,
a canonical coordinate system is the polar coordinate system, $\cc (\mathbf{x}) = (\tan^{-1}(x_2/x_1), \sqrt{x_1^2 + x_2^2})$.

\minihead{Example 2 (Horizontal Dilation)} For $T_\theta\mathbf{x} = (x_1e^\theta, x_2)$,
a canonical coordinate system is $\cc(\mathbf{x}) = \left(\log x_1, x_2\right)$.


\minihead{Reduction to Translation}  
The key property of canonical coordinates is their ability to adapt translation self-consistency to other transformation groups.
Formally, this is captured in the following result (we defer the straightforward proof to the Appendix):

\begin{proposition}
Let $f:\Phi \rightarrow \mathbb{R}^k$ be self-consistent with respect to translation and let $\rho$ be a canonical coordinate system with respect to a transformation group $G$ parameterized by $\theta\in\mathbb{R}^k$.
Then $f_\rho(\phi)\coloneqq f(\phi\circ\rho^{-1})$ is self-consistent with respect to $G$.
\end{proposition}
\vspace{-4pt}
Given a canonical coordinate system $\cc$ for a group $G$, we can thus immediately achieve self-consistency with respect to $G$ by first performing a change of coordinates into $\cc$, and then applying a function that is self-consistent with respect to translation.

\subsection{Equivariant Transformer Layers}
\label{sec:method-et}

\begin{figure}
\centering
\begin{tikzpicture}
\node[inner sep=0] (input) at (0, 0) {\includegraphics[width=2.5em]{figures/digit-90.png}};
\node[rectangle, draw, minimum width=1.5em] (f) at (2, -1.5) {$f$};
\node[] (pose) at (3.25, -1.5) {$\hat\theta$};
\node[rectangle, draw, minimum width=2.2em, minimum height=2.2em] (T) at (4.1, 0) {$T_{\hat\pose}^{-1}$};
\node[inner sep=0] (output) at (5.6, 0) {\includegraphics[width=2.5em]{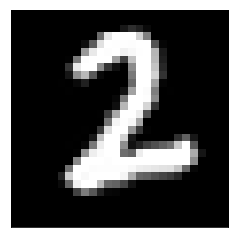}};

\draw [->] (input) to node[above]{} (T);
\draw [->] (T) to node[above]{} (output);
\draw [->] (f) to node[above]{} (pose);
\draw [->] (pose.east) -| (T.south);
\draw [->] (input.east) --++(0:6mm)|- (f);
\end{tikzpicture}

{\small(a) Spatial Transformer (ST)}
\vspace{1em}

\begin{tikzpicture}
\node[inner sep=0] (input) at (-0.25, 0) {\includegraphics[width=2.5em]{figures/digit-90.png}};
\node[rectangle, draw, minimum width=1.5em] (xi) at (1.25, -1.5) {$\cc$};
\node[inner sep=0] (polar) at (2.5, -1.5) {\includegraphics[width=2.5em]{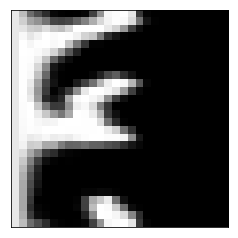}};
\node[rectangle, draw, minimum width=1.5em] (f) at (3.75, -1.5) {$f$};
\node[] (pose) at (4.75, -1.5) {$\hat\theta$};
\node[rectangle, draw, minimum width=2.2em, minimum height=2.2em] (T) at (5.25, 0) {$T_{\hat\pose}^{-1}$};
\node[inner sep=0] (output) at (6.6, 0) {\includegraphics[width=2.5em]{figures/digit-0.png}};

\draw [->] (input) to node[above]{} (T);
\draw [->] (T) to node[above]{} (output);
\draw [->] (xi) to (polar);
\draw [->] (polar) to (f);
\draw [->] (f) to node[above]{} (pose);
\draw [->] (pose.east) -| (T.south);
\draw [->] (input.east) --++(0:4mm)|- (xi);
\end{tikzpicture}

{\small(b) Equivariant Transformer (ET)}
\caption{
\textbf{Spatial and Equivariant Transformer architectures.}
In both cases, pose parameters $\hat\pose$ estimated as a function $f$ of the input image are used to apply an inverse transformation to the image.
The ET predicts $\hat\pose$ in a self-consistent manner using a canonical coordinate system $\cc$.
}
\label{fig:st-architecture}
\vspace{-4pt}
\end{figure}

Our proposed Equivariant Transformer layer leverages canonical coordinates to incorporate prior knowledge on the invariances of a domain into the network architecture:

\noindent\fbox{
  \parbox{0.95\linewidth}{
An \emph{Equivariant Transformer} (ET) layer $\et_{G, \cc} : \Img \rightarrow \Img$ for the group $G$ with canonical coordinates $\cc$ is defined as:
\begin{equation}
 \et_{G, \cc}(\img) \coloneqq T^{-1}_{f_\cc (\img)} \img
\end{equation}
where the self-consistent pose predictor $f_\cc$ is a CNN whose input is represented using the coordinates $\cc$.
  }
}

The ET layer is an image-to-image mapping that applies the inverse transformation of the predicted input pose, where the pose prediction is performed using a network that satisfies self-consistency with respect to a pre-defined group $G$.
A standard Spatial Transformer layer can be viewed as an ET where $\cc$ is simply the identity map.
Like the ST, the ET layer is differentiable with respect to both its parameters and its input; thus, it is easily incorporated as a layer in existing CNN architectures.
We summarize the computation encapsulated in the ET layer in Figure~\ref{fig:st-architecture}.

\minihead{Local Invariance}
Unlike ST layers, ET layers are endowed with a form of local transformation invariance: for any input image $\img$, we have that $L_{G,\rho}(\phi) = L_{G,\rho}(T_\theta \phi)$ for all $T_\theta \in G$.
In other words, an ET layer collapses the orbit generated by the group action on an image to a single, ``canonical'' point.
This property follows directly from the self-consistency of the pose predictor with respect to the group $G$.
Importantly, local invariance holds for \emph{any} setting of the parameters of the ET layer; thus, ETs are equipped with a strong inductive bias towards invariance with respect to the transformation group $G$.

\minihead{Implementing Self-Consistency} 
We implement translation self-consistency in $f$ by first predicting a spatial distribution by passing a 2D CNN feature map through a softmax function, and then outputting the coordinates of the centroid of this distribution.
By the translation equivariance of CNNs, a shift in the CNN input results in a corresponding shift in the predicted spatial distribution, and hence the location of the centroid. 
We rescale the centroid coordinates to match the scale of the input coordinate grid.





\subsection{Constructing Canonical Coordinates (Algorithm~\ref{alg:coords})}
\label{sec:method-deriving}

In order to construct an ET layer, we derive a canonical coordinate system for the target transformation.
Canonical coordinate systems exist for all one-parameter Lie groups~(\citealp{segman1992canonical}; Theorem~1).
For Lie groups with more than one parameter, canonical coordinates exist for \emph{Abelian} groups of dimension $k\leq d$: that is, groups whose transformations are commutative.

Here, we summarize the procedure described in \citet{segman1992canonical}.
For clarity of exposition, we will focus on Lie groups representing transformations on $\mathbb{R}^2$ with one parameter $\theta\in\mathbb{R}$.
This corresponds to the practically useful case of one-parameter deformations of 2D images.
In this setting, condition (\ref{eq:cc-condition}) reduces to:
\begin{equation*}
  \cc(T_\theta\mathbf{x}) = \cc(\mathbf{x}) + \theta \mathbf{e}_1.
\end{equation*}
Taking the derivative with respect to $\theta$, we can see that it suffices for $\cc$ to satisfy the following first-order PDEs:
\begin{align}
  \left(\frac{\partial (T_\theta \mathbf{x})_1}{\partial \theta}\bigg\rvert_{\theta=0} \frac{\partial}{\partial x_1} + \frac{\partial (T_\theta \mathbf{x})_2}{\partial \theta}\bigg\rvert_{\theta=0} \frac{\partial}{\partial x_2} \right) \cc_1(\mathbf{x}) = 1, \label{eq:coord-cov} \\
  \left(\frac{\partial (T_\theta \mathbf{x})_1}{\partial \theta}\bigg\rvert_{\theta=0} \frac{\partial}{\partial x_1} + \frac{\partial (T_\theta \mathbf{x})_2}{\partial \theta}\bigg\rvert_{\theta=0} \frac{\partial}{\partial x_2} \right) \cc_2(\mathbf{x}) = 0 . \label{eq:coord-inv}
\end{align}
We can solve these first-order PDEs using the method of characteristics~(\emph{e.g.}, \citealp{strauss2007partial}).
Observe that the homogeneous equation (\ref{eq:coord-inv}) admits an infinite set of solutions $\cc_2$; each solution is a different coordinate function that is invariant to the transformation $T_\theta$.
Thus, there exists a degree of freedom in choosing invariant coordinate functions; due to the finite resolution of images in practice, we recommend choosing coordinates that minimally distort the input image to mitigate the introduction of resampling artifacts.

\minihead{Example 3 (Hyperbolic Rotation)} As a concrete example, we will derive a set of canonical coordinates for hyperbolic rotation, $T_\theta \mathbf{x} = (x_1 e^\theta, x_2 e^{-\theta})$.
This is a ``squeeze'' distortion that dilates an image along one axis and compresses it along the other.
We obtain the following PDEs:
\begin{align*}
(x_1 \partial/\partial{x_1} - x_2 \partial/\partial{x_2})\cc_1(\mathbf{x}) &= 1, \\
(x_1 \partial/\partial{x_1} - x_2 \partial/\partial{x_2})\cc_2(\mathbf{x}) &= 0.
\end{align*}
In the first quadrant, the solution to the inhomogeneous equation is $\cc_1(\mathbf{x}) = \log \sqrt{x_1/x_2} + c_1$, where $c_1$ is an arbitrary constant, and the solution to the homogeneous equation is $\cc_2(\mathbf{x}) = h(x_1x_2)$, where $h$ is an arbitrary differentiable function in one variable (the choice $h(z) = \sqrt{z}$ is known as the hyperbolic coordinate system).
These coordinates can be defined analogously for the remaining quadrants to yield a representation of the entire image plane, excluding the lines $x_1=0$ and $x_2=0$.

\begin{algorithm}[tb]
   \caption{Constructing a canonical coordinate system}
   \label{alg:coords}
\begin{algorithmic}
   \STATE {\bfseries Input:} Transformation group $\{T_\theta\}$
   \STATE {\bfseries Output:} Canonical coordinates $\cc(\mathbf{x})$
   \STATE $v_i(\mathbf{x}) \gets (\partial(T_\theta \mathbf{x})_i / \partial\theta)\rvert_{\theta=0}, \quad i=1,2$
   \STATE $D_\mathbf{x} \gets (v_1(\mathbf{x}) \partial/\partial x_1 + v_2(\mathbf{x}) \partial/\partial x_2)$
   \STATE $\cc_1(\mathbf{x})\gets$ a solution of $D_\mathbf{x}\cc_1(\mathbf{x}) = 1$
   \STATE $\cc_2(\mathbf{x})\gets$ a solution of $D_\mathbf{x}\cc_2(\mathbf{x}) = 0$
   \STATE Return $\cc(\mathbf{x}) = (\cc_1(\mathbf{x}), \cc_2(\mathbf{x}))$
\end{algorithmic}
\end{algorithm}

\subsection{Compositions of Transformations}
\label{sec:method-composition}

A single transformation group with one parameter is typically insufficient to capture the full range of variation in object pose in natural images.
For example, an important transformation group in practice is the 8-parameter projective linear group $\mathrm{PGL}(3,\mathbb{R})$ that represents perspective transformations in 3D space.

In the special case of two-parameter Abelian Lie groups, we can construct canonical coordinates that yield self-consistency simultaneously for both parameters~(\citealp{segman1992canonical}; Theorem 1).
For example, log-polar coordinates are canonical for both rotation and dilation.
However, for transformations on $\mathbb{R}^d$, a canonical coordinate system can only satisfy condition (\ref{eq:cc-condition}) for up to $d$ parameters.
Thus, a single canonical coordinate system is insufficient for higher-dimensional transformation groups on $\mathbb{R}^2$ such as $\mathrm{PGL}(3,\mathbb{R})$.

\minihead{Stacked ETs}  Since we cannot always achieve simultaneous self-consistency with respect to all the parameters of the transformation group, we instead adopt the heuristic approach of using a sequence of ET layers, each of which implements self-consistency with respect to a subgroup of the full transformation group.
Intuitively, each ET layer aims to remove the effect of its corresponding subgroup.



Specifically, let $T_\pose$ be a $k$-parameter transformation that admits a decomposition into one-parameter transformations:
\begin{equation*}
T_\pose = T^{(1)}_{\theta'_1} \circ T^{(2)}_{\theta'_{2}} \circ \dots \circ T^{(k)}_{\theta'_{k}},
\end{equation*}
where $\theta'_i \in \mathbb{R}$.
For example, in the case of $\mathrm{PGL}(3,\mathbb{R})$, we can decompose an arbitrary transformation into a composition of one-parameter translation, dilation, rotation, shear, and perspective transformations.  
We then apply a sequence of ET layers in the reverse order of the transformations:
\begin{equation*}
\et(\phi) = \et_{G^{(k)}, \cc^{(k)}} \circ \et_{G^{(k-1)}, \cc^{(k-1)}} \circ \dots \circ \et_{G^{(1)}, \cc^{(1)}}(\phi),
\end{equation*}
where $\cc^{(i)}$ are canonical coordinates for each one-parameter subgroup $G^{(i)}$.

While we can no longer guarantee self-consistency for a composition of ET layers, we show empirically~(\S\ref{sec:experiments}) that this stacking heuristic works well in practice for transformation groups in several parameters.

\subsection{Implementation}
\label{sec:method-implementation}

Here we highlight particularly salient details of our implementation of ETs. 
Our PyTorch implementation is available at \url{github.com/stanford-futuredata/equivariant-transformers}.

\minihead{Change of Coordinates} We implement coordinate transformations by resampling the input image over a rectangular grid in the new coordinate system.
This grid consists of rows and columns that are equally spaced in the intervals $[u_1^\mathrm{min}, u_1^\mathrm{max}]$ and $[u_2^\mathrm{min}, u_2^\mathrm{max}]$, where the limits of these intervals are chosen to achieve good coverage of the input image.
These points $\mathbf{u}$ in the canonical coordinate system define a set of sampling points $\cc^{-1}(\mathbf{u})$ in Cartesian coordinates.
We use bilinear interpolation for points that do not coincide with pixel locations in the original image, as is typical with ST layers~\citep{jaderberg2015spatial}.

\minihead{Avoiding Resampling} When using multiple ET layers, iterated resampling of the input image will degrade image quality and amplify the effect of interpolation artifacts.
In our implementation, we circumvent this issue by resampling the image lazily.
More specifically, let $\phi^{(i)}$ denote the image obtained after $i$ transformations, where $\phi^{(0)}$ is the original input image.
At each iteration $i$, we represent $\phi^{(i)}$ implicitly using the sampling grid $\mathcal{G}_i \coloneqq \left(T^{(1)}_{\hat\theta_1} \circ \dots \circ T^{(i)}_{\hat\theta_i}\right) \mathcal{G}_0$, where $\mathcal{G}_0$ represents the Cartesian grid over the original input.
We materialize $\phi^{(i)}$ (under the appropriate canonical coordinates) in order to predict $\hat\theta_{i+1}$.
By appending the next predicted transformation $T^{(i+1)}_{\hat\theta_{i+1}}$ to the transformation stack, we thus obtain the subsequent sampling grid, $\mathcal{G}_{i+1}$.






\section{Experiments}
\label{sec:experiments}

We evaluate ETs on two image classification datasets: an MNIST variant where the digits are distorted under random projective transformations~(\S\ref{sec:experiment-mnist}), and the real-world Street View House Numbers (SVHN) dataset~(\S\ref{sec:experiment-svhn}).
Using projectively-transformed MNIST data, we evaluate the performance of ETs relative to STs in a setting where images are deformed by a known transformation group in several parameters.
The SVHN task evaluates the utility of ET layers when used in combination with modern CNN architectures in a realistic image classification task.
In both cases, we validate the sample efficiency benefits conferred by ETs relative to standard STs and baseline CNN architectures.\footnote{In the Appendix, we report additional experimental results on robustness to transformations not seen at training time.}


\subsection{Projective MNIST}
\label{sec:experiment-mnist}

\begin{figure}
\centering
\includegraphics[width=0.95\linewidth]{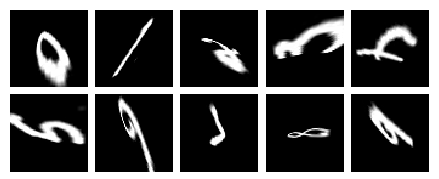}
\vspace{-4pt}
\caption{
\textbf{Projective MNIST.}
Examples of transformed digits from each class (first row: 0--4, second row: 5--9).
Each base MNIST image is transformed using a transformation sampled from a 6-parameter group (\emph{i.e.}, $\mathrm{PGL}(3, \mathbb{R})$ without translation).
}
\label{fig:mnist-proj}
\vspace{-6pt}
\end{figure}

We introduce the Projective MNIST dataset, a variant of the MNIST dataset where the digits are distorted using randomly sampled projective transformations: namely rotation, shear, $x$- and $y$-dilation, and $x$- and $y$-perspective transformations~(\emph{i.e.}, 6 pose parameters in total).
The Projective MNIST training set contains 10,000 base images sampled without replacement from the MNIST training set.
Each image is resized to $64\times64$ and transformed using an independently-sampled set of pose parameters.

We also generated three larger versions of the dataset for the purpose of controlled evaluation of the effect of (idealized) data augmentation: these additional datasets respectively contain 2, 4, and 8 copies of the base MNIST images, each transformed under different sets of parameters.

Unlike other MNIST variants such as Rotated MNIST~\citep{larochelle2007empirical}, MNIST-RTS~\citep{jaderberg2015spatial}, and SIM2MNIST~\citep{esteves2018polar}, our Projective MNIST dataset incorporates higher-dimensional combinations of transformations, including projective transformations not considered in prior work (\emph{e.g.}, perspective transforms).
We provide further details on the construction of the dataset in the Appendix.


\minihead{Network Architectures} We used a CNN architecture based on the Z2CNN from \citet{cohen2016group}, with 7 layers of $3\times 3$ convolutions with 32 channels, batch normalization after convolutional layers, and dropout after the 3rd and 6th layers.
In addition to this baseline ``Cartesian'' CNN, we also evaluated a more rotation- and dilation-robust network where the inputs are first transformed to log-polar coordinates~\citep{henriques2017warped,esteves2018polar}.

We introduce a sequence of transformer layers before the log-polar coordinate transformation to handle the remaining geometric transformations applied to the input.
For both the baseline STs and ETs, we apply a sequence of transformer layers, with each layer predicting a single pose parameter.
The pose predictor networks in both cases are 3-layer CNNs with 32 channels in each layer.
We selected the transformation order, dropout rate, and learning rate schedule based on validation accuracy~(see the Appendix for details).

\begin{table}[t]
\caption{
\textbf{Classification error rates on Projective MNIST~(\S\ref{sec:experiment-mnist}).}
All methods use the same CNN architecture for classification and differ in the transformations applied to the input images.
We train on up to 8 sampled transformations for each base MNIST image.
LP: log-polar coordinates; $\mathrm{sh}_x$: $x$-shear; $\mathrm{hr}$: hyperbolic rotation; $\mathrm{p}_x$: $x$-perspective; $\mathrm{p}_y$: $y$-perspective.
\vspace{0.25em}
}
\label{tab:experiments-mnist-proj}

\centering
\resizebox{\columnwidth}{!}{%
\begin{tabular}{llrrrr}
\toprule 
\multirow{2}{*}{Method} & \multirow{2}{*}{Transformations} & \multicolumn{4}{c}{\# sampled transformations} \\
\cmidrule(lr){3-6}
& & 1 & 2& 4& 8 \\
\midrule
Cartesian & - & 11.91 & 9.67 & 7.64 & 6.93 \\
Log-polar & - & 6.55 & 5.05 & 4.48 & 3.83 \\ \midrule
ST-LP  & $\mathrm{sh}_x$ & 5.77 & 4.27 & 3.97 & 3.47\\
ST-LP  & $\mathrm{sh}_x \,\Arrow{2mm}\, \mathrm{hr}$ & 4.92 & 3.87 & 3.22 & 3.03 \\
ST-LP*  & $\mathrm{sh}_x \,\Arrow{2mm}\, \mathrm{hr} \,\Arrow{2mm}\, \mathrm{p}_x \,\Arrow{2mm}\, \mathrm{p}_y$ & -- & -- & -- & -- \\
\midrule
ET-LP  & $\mathrm{sh}_x$ & 5.48 & 4.67 & 3.63 & 3.21 \\
ET-LP  & $\mathrm{sh}_x \,\Arrow{2mm}\, \mathrm{hr}$ & 4.18 & 3.17 & 2.96 & 2.62 \\
ET-LP  & $\mathrm{sh}_x \,\Arrow{2mm}\, \mathrm{hr} \,\Arrow{2mm}\, \mathrm{p}_x \,\Arrow{2mm}\, \mathrm{p}_y$ &  \textbf{3.76} & \textbf{3.11} & \textbf{2.80} & \textbf{2.60} \\
\bottomrule
\end{tabular}
}
\begin{flushleft}
{\fontsize{8}{9}\selectfont *We omit this configuration due to training instability.}
\end{flushleft}
\vspace{-6pt}
\end{table}

\minihead{Classification Accuracy (Table~\ref{tab:experiments-mnist-proj})} We find that the ET layers consistently improve on test error rate over both the log-polar and ST baselines.
By accounting for additional transformations, the ET improves on the error rate of the baseline log-polar CNN by 2.79\%---a relative improvement of 43\%---when trained on a single pose per prototype.
Note that we omit the ST baseline with the full transformation sequence due to training instability, despite more extensive hyperparameter tuning than the ET.
We find that all methods improve from augmentation with additional poses, with the ET retaining its advantage but at a reduced margin. 

\begin{figure}
\centering
\includegraphics[width=0.9\linewidth]{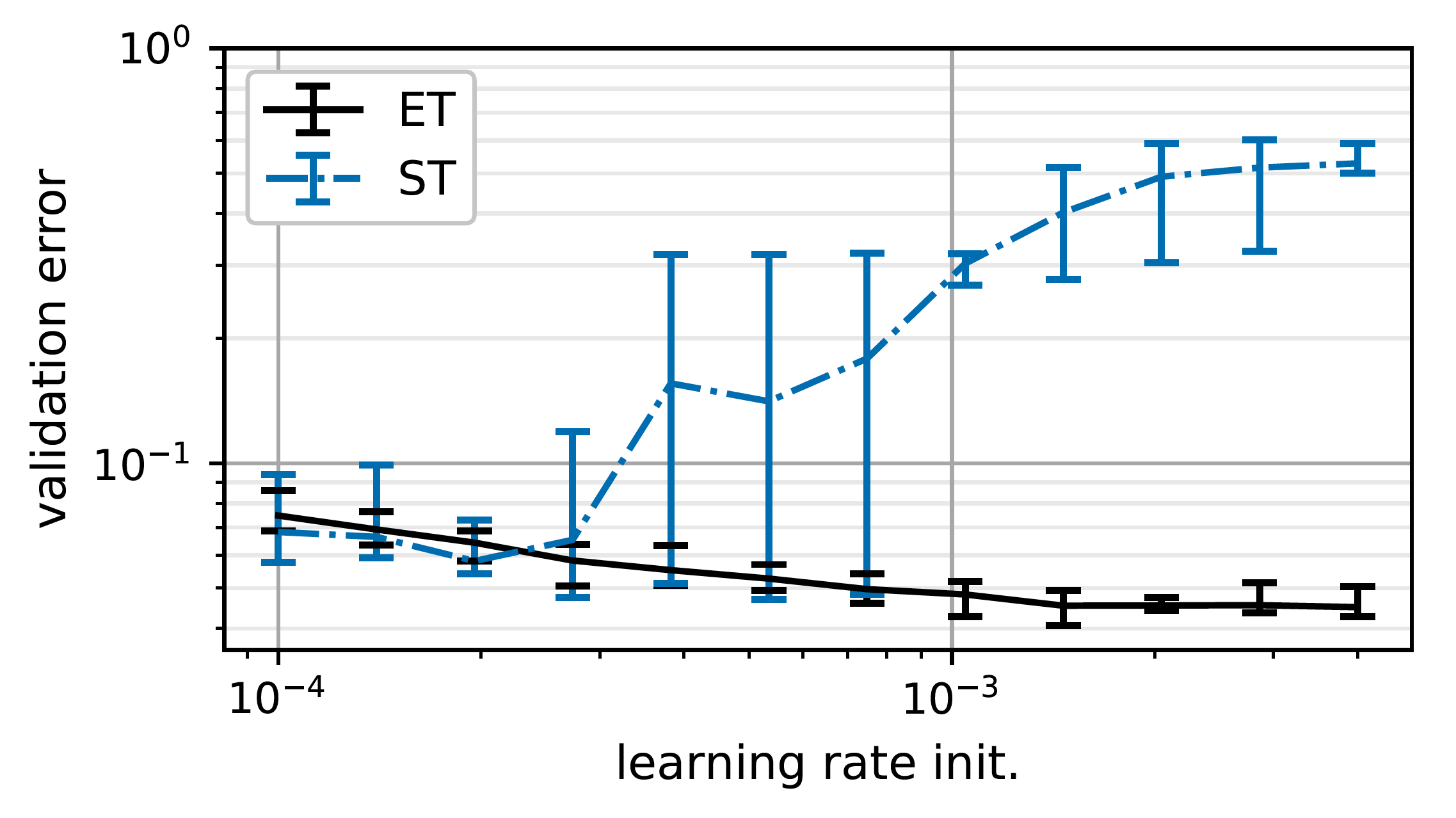}
\vspace{-6pt}
\caption{
\textbf{Sensitivity to initial learning rate.}
For each learning rate setting, we plot the minimum, mean, and maximum validation error rates over 10 runs for networks trained with ETs and STs.
The predicted transformations are $x$-shear and hyperbolic rotation.
We find that ETs are significantly more robust than STs to the learning rate hyperparameter.
}
\vspace{-4pt}
\label{fig:hyperparameter-sensitivity}
\end{figure}

\minihead{Hyperparameter Sensitivity~(Figure~\ref{fig:hyperparameter-sensitivity})} 
We compared the sensitivity of ET and ST networks to the initial learning rate by comparing validation error when training with learning rate values ranging from $1\times10^{-4}$ to $4\times10^{-3}$.
For each setting, we trained 10 networks with independent random initializations on Projective MNIST with 10,000 examples, computing the validation error after each epoch and recording the minimum observed error in each run.
We find that STs were significantly more sensitive to learning rate than ET, with far higher variance in error rate between runs.
This suggests that the self-consistency constraint imposed on ETs helps improve the training-time stability of networks augmented with transformer layers.  



\subsection{Street View House Numbers (SVHN)}
\label{sec:experiment-svhn}

The goal of the single-digit classification task of the SVHN dataset~\citep{netzer2011reading} is to classify the digit in the center of $32\times 32$ RGB images of house numbers.
SVHN is well-suited to evaluating the effect of transformer layers since there is a natural range of geometric variation in the data due to differences in camera position---unlike Projective MNIST, we do not artificially apply further transformations to the data.
The training set consists of 73,257 examples; we use a randomly-chosen subset of 5,000 examples for validation and use the remaining 68,257 examples for training.
In order to evaluate the data efficiency of each method, we also trained models using smaller subsets of 10,000 and 20,000 examples.
The dataset also includes 531,131 additional images that can be used as extra training data; we thus additionally evaluate our methods on the concatenation of this set and the training set.

\minihead{Network Architectures} We use 10-, 18-, and 34-layer ResNet architectures~\citep{he2016deep} as baseline networks.
Each transformer layer uses a 3-layer CNN with 32 channels per layer for pose prediction.
We applied $x$- and $y$-translation, rotation, and $x$- and $y$-scaling to the input images: these were selected from among the subgroups of the projective group using the validation set.


\begin{table}[t]
\caption{
\textbf{Classification error rates on SVHN~(\S\ref{sec:experiment-svhn}).}
For both STs and ETs, we used the following transformations: $x$- and $y$-translation, rotation, and $x$-scaling.
Error rates are each averaged over 3 runs.
ETs achieve the largest accuracy gains relative to STs and the baseline CNNs in the limited data regime.
\vspace{0.25em}
}
\label{tab:svhn-results}
\centering
\resizebox{\columnwidth}{!}{%
\begin{tabular}{llrrrr}
\toprule
\multirow{2}{*}{Network} & \multirow{2}{*}{Transformer} & \multicolumn{4}{c}{\# training examples} \\
\cmidrule(lr){3-6}
       & & 10k & 20k & 68k & 600k \\
\midrule
ResNet-10 & None & 9.83 & 7.90 & 5.35 & 2.96 \\
  & Spatial & 9.80 & 7.66 & 4.96 & 2.92 \\
  & Equivariant & \textbf{8.24} & \textbf{6.71} & \textbf{4.84} & \textbf{2.70} \\ \midrule
ResNet-18  & None & 9.23 & 7.31 & 4.81 & 2.76 \\
  & Spatial & 9.10 & 7.17 & 4.51 & 2.70 \\
  & Equivariant &  \textbf{7.81} & \textbf{6.37} &	\textbf{4.50} & \textbf{2.57} \\ \midrule
ResNet-34  & None & 8.73	& 7.05 & 4.67 & 2.53 \\
  & Spatial &  8.60 & 6.91 & 4.37 & 2.66 \\
  & Equivariant & \textbf{7.72} & \textbf{5.98} & \textbf{4.23} & \textbf{2.47} \\
\bottomrule
\end{tabular}
}
\vspace{-12pt}
\end{table}

\vspace{-4pt}
\minihead{Results (Table~\ref{tab:svhn-results})} We find that ETs improve on the error rate achieved by both STs and the baseline ResNets, with the largest gains seen in the limited data regime: with 10,000 examples, ETs improve on the error rates of the baseline CNNs and ST-augmented CNNs by 0.9--1.6\%, or a relative improvement of 10--16\%.
We see smaller gains when more training data is available: the relative improvement between ETs and the baseline CNNs is 11--13\% with 20,000 examples, and 6.4--9.5\% with 68,257 examples.

When data is limited, we find that a simpler classifier where prior knowledge on geometric invariances has been encoded using ETs can outperform more complex classifiers that are not equipped with this additional structure.
In particular, when trained on 10,000 examples, a ResNet-10 classifier with ET layers achieves lower error than the baseline ResNet-34 classifier.
The baseline ResNet-34 has over 5.3M parameters; in contrast, the ResNet-10 has 1.2M parameters, with the ET layers adding only 31k parameters in total.
The ET-augmented ResNet-10 therefore achieves improved error rate with an architecture that incurs less memory and computational cost than a ResNet-34.


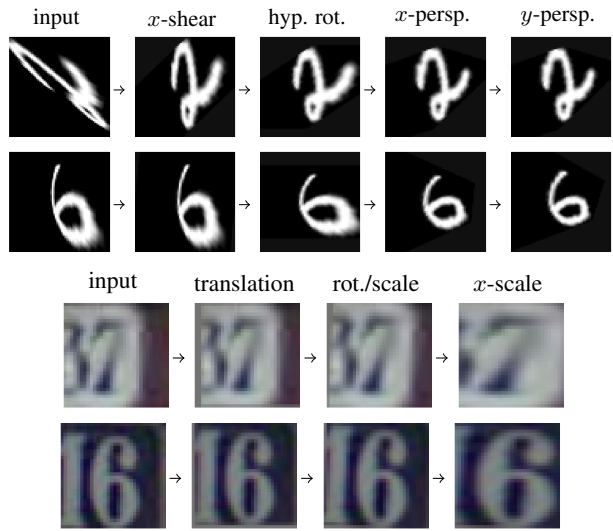
\begin{figure}

\centering
\resizebox{\columnwidth}{!}{%
\begin{tikzpicture}
\node[inner sep=0] (1) at (0, 0) {\includegraphics[width=4.5em]{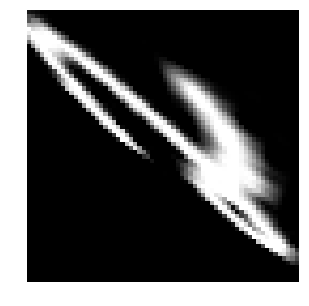}};
\node[inner sep=0] (2) at (1.75, 0) {\includegraphics[width=4.5em]{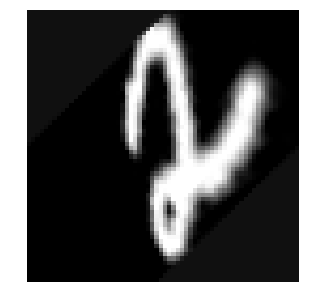}};
\node[inner sep=0] (3) at (3.5, 0) {\includegraphics[width=4.5em]{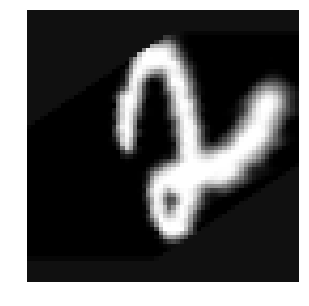}};
\node[inner sep=0] (4) at (5.25, 0) {\includegraphics[width=4.5em]{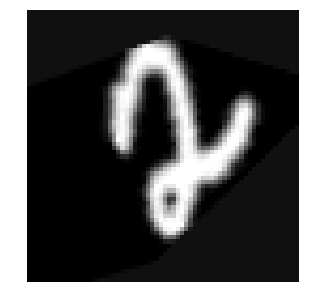}};
\node[inner sep=0] (5) at (7, 0) {\includegraphics[width=4.5em]{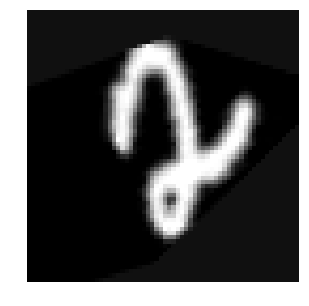}};

\node[] (input) at (0, 1) {\footnotesize input};
\node[] (a) at (1.75, 1) {\footnotesize $x$-shear};
\node[] (b) at (3.5, 1) {\footnotesize hyp. rot.};
\node[] (c) at (5.25, 1) {\footnotesize $x$-persp.};
\node[] (d) at (7, 1) {\footnotesize $y$-persp.};

\draw [->] (1) to node[above]{} (2);
\draw [->] (2) to node[above]{} (3);
\draw [->] (3) to node[above]{} (4);
\draw [->] (4) to node[above]{} (5);
\end{tikzpicture}
}

%

\resizebox{\columnwidth}{!}{%
\begin{tikzpicture}
\node[inner sep=0] (1) at (0, 0) {\includegraphics[width=4.5em]{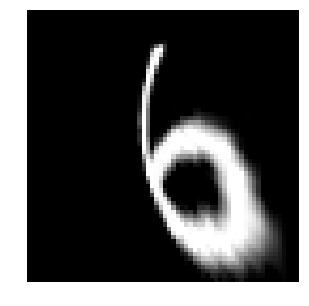}};
\node[inner sep=0] (2) at (1.75, 0) {\includegraphics[width=4.5em]{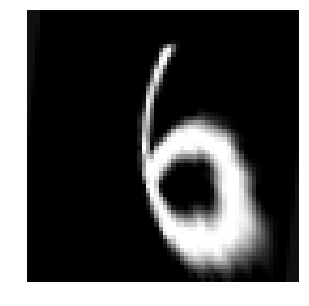}};
\node[inner sep=0] (3) at (3.5, 0) {\includegraphics[width=4.5em]{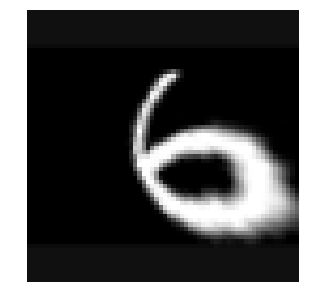}};
\node[inner sep=0] (4) at (5.25, 0) {\includegraphics[width=4.5em]{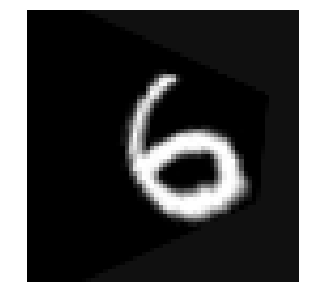}};
\node[inner sep=0] (5) at (7, 0) {\includegraphics[width=4.5em]{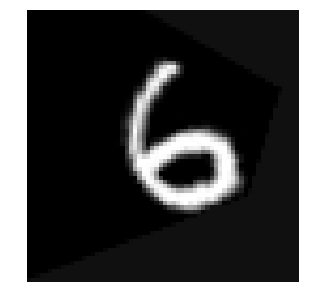}};

\draw [->] (1) to node[above]{} (2);
\draw [->] (2) to node[above]{} (3);
\draw [->] (3) to node[above]{} (4);
\draw [->] (4) to node[above]{} (5);
\end{tikzpicture}
}

%

\begin{tikzpicture}
\node[inner sep=0] (1) at (0, 0) {\includegraphics[width=4.5em]{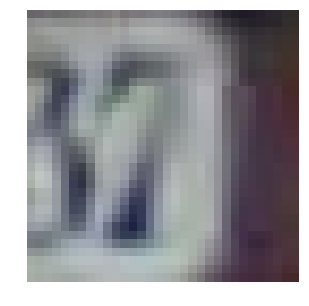}};
\node[inner sep=0] (2) at (1.75, 0) {\includegraphics[width=4.5em]{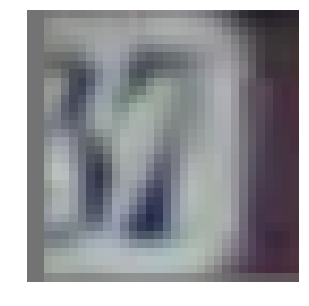}};
\node[inner sep=0] (3) at (3.5, 0) {\includegraphics[width=4.5em]{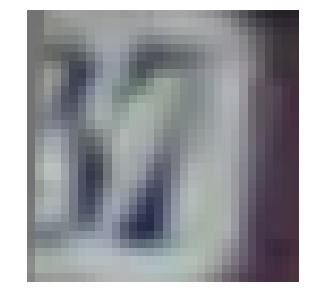}};
\node[inner sep=0] (4) at (5.25, 0) {\includegraphics[width=4.5em]{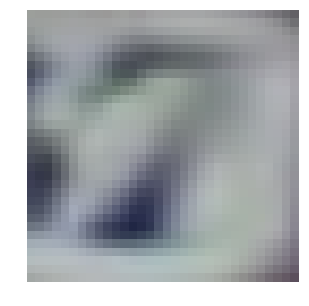}};

\node[] (input) at (0, 1) {\footnotesize input};
\node[] (a) at (1.75, 1) {\footnotesize translation};
\node[] (b) at (3.5, 1) {\footnotesize rot./scale};
\node[] (c) at (5.25, 1) {\footnotesize $x$-scale};

\draw [->] (1) to node[above]{} (2);
\draw [->] (2) to node[above]{} (3);
\draw [->] (3) to node[above]{} (4);
\end{tikzpicture}

\begin{tikzpicture}
\node[inner sep=0] (1) at (0, 0) {\includegraphics[width=4.5em]{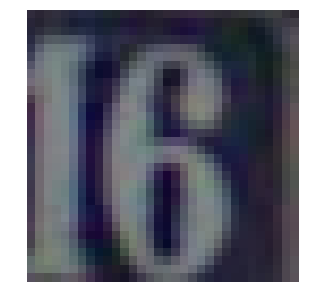}};
\node[inner sep=0] (2) at (1.75, 0) {\includegraphics[width=4.5em]{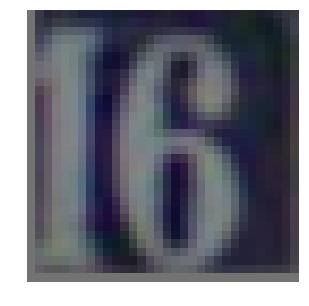}};
\node[inner sep=0] (3) at (3.5, 0) {\includegraphics[width=4.5em]{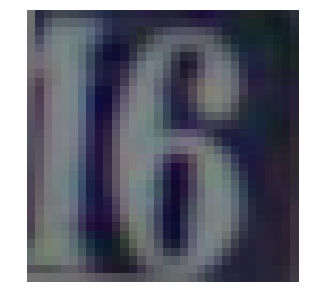}};
\node[inner sep=0] (4) at (5.25, 0) {\includegraphics[width=4.5em]{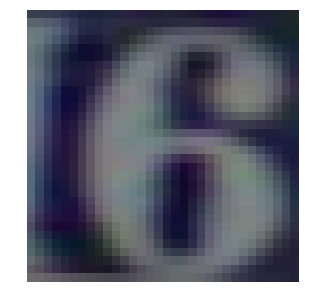}};

\draw [->] (1) to node[above]{} (2);
\draw [->] (2) to node[above]{} (3);
\draw [->] (3) to node[above]{} (4);
\end{tikzpicture}
\vspace{-1.1em}
\caption{\textbf{Predicted transformations.} 
On Projective MNIST (top), ETs reverse the effect of distortions such as shear and perspective, despite being provided no direct supervision on pose parameters (the final images remain rotated and scaled since the classification CNN operates over their log-polar representation).
On SVHN (bottom), the final $x$-scale transformation has a cropping effect that removes distractor digits. 
}
\label{fig:visualization}
\vspace{-8pt}
\end{figure}

\vspace{-6pt}

\section{Discussion and Conclusion}
\label{sec:discussion}

\vspace{-8pt}
\minihead{Limitations of ETs}
The self-consistency guarantee of ETs can fail due to boundary effects that occur when image content is cropped after a transformation.
This issue can be mitigated by padding the input such that the transformed image does not fall ``out of frame''.
Even without a strict self-consistency guarantee, we still observe gains when ET layers are used in practice~(\emph{e.g.}, in our SVHN experiments).

As discussed in \S\ref{sec:method-composition}, the method of stacking ET layers is ultimately a heuristic approach as it does not guarantee self-consistency with respect to the full transformation group.
Moreover, higher-dimensional groups require the use of long sequences of ET layers, resulting in high computational cost.
In such cases, we could employ a hybrid approach where ``difficult'' subgroups are handled by ET layers, while the remaining degrees of freedom are handled by a standard ST layer.
In general, enforcing equivariance guarantees for higher-dimensional transformation groups in a computationally scalable fashion remains an open problem.

In contrast to the use of prior knowledge on transformation invariances in this work, there is a separate line of research that concerns learning various classes of transformations from data~\citep{hashimoto2017unsupervised,thomas2018learning}.
Extending ETs to these more flexible notions of invariance may prove to be an interesting direction for future work.

\vspace{-6pt}
\minihead{Conclusion}
We proposed a neural network layer that builds in prior knowledge on the continuous transformation invariances of its input domain.
By encapsulating equivariant functions within an image-to-image mapping, ETs expose a convenient interface for flexible composition of layers tailored to different transformation groups.
Empirically, we demonstrated that ETs improve the sample efficiency of CNNs on image classification tasks with latent transformation parameters.
Using libraries of ET layers, practitioners are able to quickly experiment with multiple combinations of transformations to realize gains in predictive accuracy, particularly in domains where labeled data is scarce.

\section*{Acknowledgements}

We thank Pratiksha Thaker, Kexin Rong, and our anonymous reviewers for their valuable feedback on earlier versions of this manuscript.
This research was supported in part by affiliate members and other supporters of the Stanford DAWN project---Ant Financial, Facebook, Google, Intel, Microsoft, NEC, SAP, Teradata, and VMware---as well as Toyota Research Institute, Keysight Technologies, Northrop Grumman, Hitachi, NSF awards AF-1813049 and CCF-1704417, an ONR Young Investigator Award N00014-18-1-2295, and Department of Energy award DE-SC0019205.

\bibliography{references}

\begin{thebibliography}{34}
\providecommand{\natexlab}[1]{#1}
\providecommand{\url}[1]{\texttt{#1}}
\expandafter\ifx\csname urlstyle\endcsname\relax
  \providecommand{\doi}[1]{doi: #1}\else
  \providecommand{\doi}{doi: \begingroup \urlstyle{rm}\Url}\fi

\bibitem[Amari(1978)]{amari1978feature}
Amari, S.
\newblock {Feature Spaces which Admit and Detect Invariant Signal
  Transformations}.
\newblock In \emph{International Joint Conference on Pattern Recognition},
  1978.

\bibitem[Amit et~al.(1991)Amit, Grenander, and Piccioni]{amit1991structural}
Amit, Y., Grenander, U., and Piccioni, M.
\newblock Structural image restoration through deformable templates.
\newblock \emph{Journal of the American Statistical Association}, 1991.

\bibitem[Cohen \& Welling(2016)Cohen and Welling]{cohen2016group}
Cohen, T.~S. and Welling, M.
\newblock {Group Equivariant Convolutional Networks}.
\newblock In \emph{International Conference on Machine Learning}, 2016.

\bibitem[Cohen et~al.(2018)Cohen, Geiger, K{\"o}hler, and
  Welling]{cohen2018spherical}
Cohen, T.~S., Geiger, M., K{\"o}hler, J., and Welling, M.
\newblock {Spherical CNNs}.
\newblock In \emph{International Conference on Learning Representations}, 2018.

\bibitem[De~Castro \& Morandi(1987)De~Castro and Morandi]{de1987registration}
De~Castro, E. and Morandi, C.
\newblock {Registration of translated and rotated images using finite Fourier
  transforms}.
\newblock \emph{IEEE Transactions on Pattern Analysis and Machine
  Intelligence}, pp.\  700--703, 1987.

\bibitem[Dieleman et~al.(2016)Dieleman, De~Fauw, and
  Kavukcuoglu]{dieleman2016exploiting}
Dieleman, S., De~Fauw, J., and Kavukcuoglu, K.
\newblock Exploiting cyclic symmetry in convolutional neural networks.
\newblock \emph{International Conference on Machine Learning}, 2016.

\bibitem[Esteves et~al.(2018)Esteves, Allen-Blanchette, Zhou, and
  Daniilidis]{esteves2018polar}
Esteves, C., Allen-Blanchette, C., Zhou, X., and Daniilidis, K.
\newblock {Polar Transformer Networks}.
\newblock In \emph{International Conference on Learning Representations}, 2018.

\bibitem[Freeman \& Adelson(1991)Freeman and Adelson]{freeman1991design}
Freeman, W.~T. and Adelson, E.~H.
\newblock The design and use of steerable filters.
\newblock \emph{IEEE Transactions on Pattern Analysis and Machine
  Intelligence}, pp.\  891--906, 1991.

\bibitem[Gens \& Domingos(2014)Gens and Domingos]{gens2014deep}
Gens, R. and Domingos, P.~M.
\newblock {Deep Symmetry Networks}.
\newblock In \emph{Advances in Neural Information Processing Systems}, 2014.

\bibitem[Hashimoto et~al.(2017)Hashimoto, Liang, and
  Duchi]{hashimoto2017unsupervised}
Hashimoto, T.~B., Liang, P.~S., and Duchi, J.~C.
\newblock Unsupervised transformation learning via convex relaxations.
\newblock In \emph{Advances in Neural Information Processing Systems}, 2017.

\bibitem[He et~al.(2016)He, Zhang, Ren, and Sun]{he2016deep}
He, K., Zhang, X., Ren, S., and Sun, J.
\newblock Deep residual learning for image recognition.
\newblock In \emph{IEEE Conf. on Computer Vision and Pattern Recognition
  (CVPR)}, 2016.

\bibitem[Henriques \& Vedaldi(2017)Henriques and Vedaldi]{henriques2017warped}
Henriques, J.~F. and Vedaldi, A.
\newblock {Warped Convolutions: Efficient Invariance to Spatial
  Transformations}.
\newblock In \emph{International Conference on Machine Learning}, 2017.

\bibitem[Jacobsen et~al.(2017)Jacobsen, De~Brabandere, and
  Smeulders]{jacobsen2017dynamic}
Jacobsen, J.-H., De~Brabandere, B., and Smeulders, A.~W.
\newblock {Dynamic Steerable Blocks in Deep Residual Networks}.
\newblock \emph{arXiv preprint arXiv:1706.00598}, 2017.

\bibitem[Jaderberg et~al.(2015)Jaderberg, Simonyan, Zisserman, and
  Kavukcuoglu]{jaderberg2015spatial}
Jaderberg, M., Simonyan, K., Zisserman, A., and Kavukcuoglu, K.
\newblock {Spatial Transformer Networks}.
\newblock In \emph{Advances in Neural Information Processing Systems}, 2015.

\bibitem[Laptev et~al.(2016)Laptev, Savinov, Buhmann, and
  Pollefeys]{laptev2016ti}
Laptev, D., Savinov, N., Buhmann, J.~M., and Pollefeys, M.
\newblock {TI-POOLING: transformation-invariant pooling for feature learning in
  convolutional neural networks}.
\newblock In \emph{IEEE Conf. on Computer Vision and Pattern Recognition
  (CVPR)}, 2016.

\bibitem[Larochelle et~al.(2007)Larochelle, Erhan, Courville, Bergstra, and
  Bengio]{larochelle2007empirical}
Larochelle, H., Erhan, D., Courville, A., Bergstra, J., and Bengio, Y.
\newblock An empirical evaluation of deep architectures on problems with many
  factors of variation.
\newblock In \emph{International Conference on Machine Learning}, 2007.

\bibitem[Lin \& Lucey(2017)Lin and Lucey]{lin2017inverse}
Lin, C.-H. and Lucey, S.
\newblock {Inverse Compositional Spatial Transformer Networks}.
\newblock \emph{IEEE Conf. on Computer Vision and Pattern Recognition (CVPR)},
  2017.

\bibitem[Lucas \& Kanade(1981)Lucas and Kanade]{lucas1981iterative}
Lucas, B.~D. and Kanade, T.
\newblock An iterative image registration technique with an application to
  stereo vision.
\newblock In \emph{International Joint Conference on Artificial intelligence},
  1981.

\bibitem[Marcos et~al.(2017)Marcos, Volpi, Komodakis, and
  Tuia]{marcos2017rotation}
Marcos, D., Volpi, M., Komodakis, N., and Tuia, D.
\newblock {Rotation Equivariant Vector Field Networks}.
\newblock In \emph{International Conference on Computer Vision}, 2017.

\bibitem[Netzer et~al.(2011)Netzer, Wang, Coates, Bissacco, Wu, and
  Ng]{netzer2011reading}
Netzer, Y., Wang, T., Coates, A., Bissacco, A., Wu, B., and Ng, A.~Y.
\newblock Reading digits in natural images with unsupervised feature learning.
\newblock In \emph{NIPS workshop on deep learning and unsupervised feature
  learning}, 2011.

\bibitem[Rawlinson et~al.(2018)Rawlinson, Ahmed, and
  Kowadlo]{rawlinson2018sparse}
Rawlinson, D., Ahmed, A., and Kowadlo, G.
\newblock Sparse unsupervised capsules generalize better.
\newblock \emph{arXiv preprint arXiv:1804.06094}, 2018.

\bibitem[Reddi et~al.(2018)Reddi, Kale, and Kumar]{reddi2018convergence}
Reddi, S.~J., Kale, S., and Kumar, S.
\newblock {On the Convergence of Adam and Beyond}.
\newblock In \emph{International Conference on Learning Representations}, 2018.

\bibitem[Reddy \& Chatterji(1996)Reddy and Chatterji]{reddy1996fft}
Reddy, B.~S. and Chatterji, B.~N.
\newblock {An FFT-based technique for translation, rotation, and
  scale-invariant image registration}.
\newblock \emph{IEEE Transactions on Image Processing}, 5\penalty0
  (8):\penalty0 1266--1271, 1996.

\bibitem[Rubinstein et~al.(1991)Rubinstein, Segman, and
  Zeevi]{rubinstein1991recognition}
Rubinstein, J., Segman, J., and Zeevi, Y.
\newblock Recognition of distorted patterns by invariance kernels.
\newblock \emph{Pattern Recognition}, 24\penalty0 (10):\penalty0 959--967,
  1991.

\bibitem[Sabour et~al.(2017)Sabour, Frosst, and Hinton]{sabour2017dynamic}
Sabour, S., Frosst, N., and Hinton, G.~E.
\newblock Dynamic routing between capsules.
\newblock In \emph{Advances in Neural Information Processing Systems}, 2017.

\bibitem[Segman et~al.(1992)Segman, Rubinstein, and Zeevi]{segman1992canonical}
Segman, J., Rubinstein, J., and Zeevi, Y.~Y.
\newblock {The canonical coordinates method for pattern deformation:
  Theoretical and computational considerations}.
\newblock \emph{IEEE Transactions on Pattern Analysis and Machine
  Intelligence}, pp.\  1171--1183, 1992.

\bibitem[Shu et~al.(2018)Shu, Sahasrabudhe, Alp~Guler, Samaras, Paragios, and
  Kokkinos]{shu2018deforming}
Shu, Z., Sahasrabudhe, M., Alp~Guler, R., Samaras, D., Paragios, N., and
  Kokkinos, I.
\newblock {Deforming Autoencoders: Unsupervised Disentangling of Shape and
  Appearance}.
\newblock In \emph{European Conference on Computer Vision (ECCV)}, 2018.

\bibitem[Strauss(2007)]{strauss2007partial}
Strauss, W.~A.
\newblock \emph{{Partial Differential Equations: An Introduction}}.
\newblock Wiley, 2007.

\bibitem[Teo(1998)]{teo1998theory}
Teo, P.~C.
\newblock \emph{{Theory and Applications of Steerable Functions}}.
\newblock PhD thesis, Stanford University, 1998.

\bibitem[Thomas et~al.(2018)Thomas, Gu, Dao, Rudra, and
  R{\'e}]{thomas2018learning}
Thomas, A., Gu, A., Dao, T., Rudra, A., and R{\'e}, C.
\newblock Learning compressed transforms with low displacement rank.
\newblock In \emph{Advances in Neural Information Processing Systems}, 2018.

\bibitem[Weiler et~al.(2018)Weiler, Hamprecht, and Storath]{weiler2018learning}
Weiler, M., Hamprecht, F.~A., and Storath, M.
\newblock {Learning Steerable Filters for Rotation Equivariant CNNs}.
\newblock In \emph{IEEE Conf. on Computer Vision and Pattern Recognition
  (CVPR)}, 2018.

\bibitem[Worrall et~al.(2017)Worrall, Garbin, Turmukhambetov, and
  Brostow]{worrall2017harmonic}
Worrall, D.~E., Garbin, S.~J., Turmukhambetov, D., and Brostow, G.~J.
\newblock {Harmonic Networks: Deep Translation and Rotation Equivariance}.
\newblock In \emph{IEEE Conf. on Computer Vision and Pattern Recognition
  (CVPR)}, 2017.

\bibitem[Yuille(1991)]{yuille1991deformable}
Yuille, A.~L.
\newblock Deformable templates for face recognition.
\newblock \emph{Journal of Cognitive Neuroscience}, 1991.

\bibitem[Zhou et~al.(2017)Zhou, Ye, Qiu, and Jiao]{zhou2017oriented}
Zhou, Y., Ye, Q., Qiu, Q., and Jiao, J.
\newblock Oriented response networks.
\newblock In \emph{IEEE Conf. on Computer Vision and Pattern Recognition
  (CVPR)}, 2017.

\end{thebibliography}
\bibliographystyle{icml2019}

\newpage
\onecolumn
\begin{appendix}
\section{Proof of Proposition 1}

\begin{proposition}
Let $f:\Phi \rightarrow \mathbb{R}^k$ be self-consistent with respect to translation and let $\rho$ be a canonical coordinate system with respect to a transformation group $G$ parameterized by $\theta\in\mathbb{R}^k$.
Then $f_\rho(\phi)\coloneqq f(\phi\circ\rho^{-1})$ is self-consistent with respect to $G$.

\begin{proof}
By the definition of $\cc$,
\begin{align*}
  (\cc\circ T_\theta \circ \cc^{-1})(\mathbf{u}) &= \cc(\cc^{-1}(\mathbf{u})) + \sum_{i=1}^k\theta_i\mathbf{e}_i \\
   	&= \mathbf{u} + \sum_{i=1}^k\theta_i\mathbf{e}_i,
\end{align*}
and therefore $(T_\theta \circ \cc^{-1})(\mathbf{u}) = \cc^{-1}\left(\mathbf{u} + \sum_{i=1}^k \theta_i\mathbf{e}_i\right)$.
By this identity and translation self-consistency of $f$,
\begin{align*}
f_\cc(T_\theta \phi) &= f((T_\theta \phi\circ \cc^{-1})(\mathbf{u})) \\
  &= f((\phi \circ T_\theta \circ \cc^{-1})(\mathbf{u}) )\\
  &= f\left((\phi \circ \cc^{-1})\left(\mathbf{u} + \sum_{i=1}^k\theta_i\mathbf{e}_i\right)\right) \\
  &= f((\phi \circ \cc^{-1})(\mathbf{u})) + \theta \\
  &= f_\cc(\phi) + \theta,
\end{align*}
where in the second line we used the definition of $T_\theta \phi$, in the third line we used the identity for $(T_\theta\circ \cc^{-1})(\mathbf{u})$, and in the fourth line we used the translation self-consistency of $f$.
This establishes self-consistency with respect to $G$.
\end{proof}
\end{proposition}

\section{Canonical Coordinate Systems}
\label{appendix:canonical}

In Table~\ref{tab:coords}, we list the set of canonical coordinates that we derived for our experiments along with their corresponding transformation groups.
As explained in the main text, these coordinates are not unique for one-parameter transformation groups: in this case, there exists a degree of freedom in specifying the complementary set of coordinates.

\begin{figure}
\centering
\includegraphics[width=0.5\textwidth]{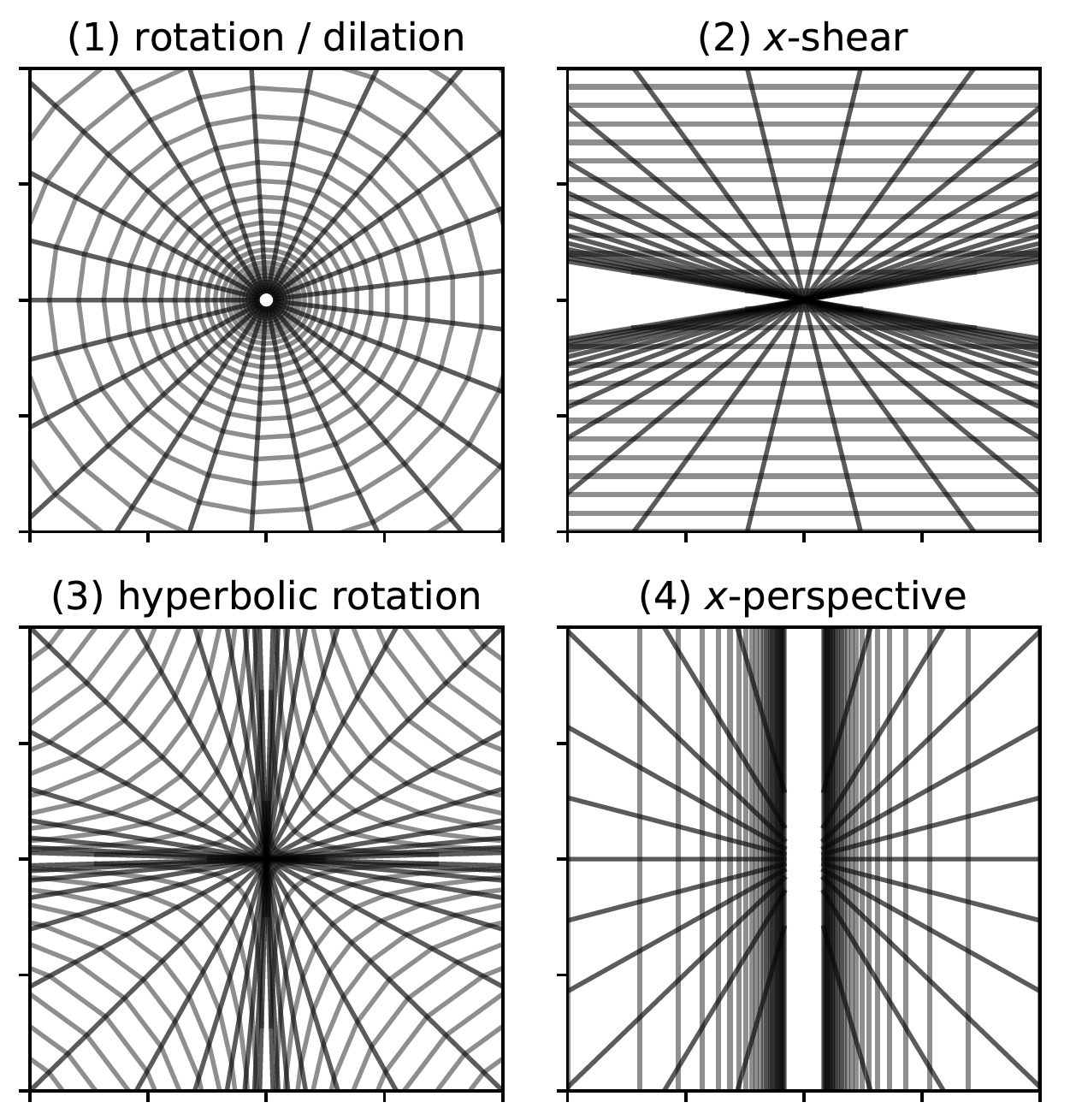}
\vspace{-1em}
\caption{
\textbf{Examples of canonical coordinate systems.}
The corresponding images of $(x, y)$ for each transformation are \textbf{(1a)}~rotation, \textbf{(1b)}~dilation, \textbf{(2)}~$x$-shear, \textbf{(3)}~hyperbolic rotation, and \textbf{(4)}~$x$-perspective.
}
\label{fig:canonical-coords}
\end{figure}

In Figure~\ref{fig:canonical-coords}, we plot some examples of canonical coordinate grids used in our experiments.

\section{Experimental Details}

\subsection{Projective MNIST}

\minihead{Dataset} To construct the Projective MNIST dataset, we sampled 10,000 images without replacement from the MNIST training set (consisting of 60,000 examples).
Each $28\times28$ base image is extended to $64\times 64$ by symmetric zero padding.
The images are then distorted using transformations sampled independently from the projective group.
The 6 pose parameters were sampled uniformly from the ranges listed in Table~\ref{tab:ranges}.
We excluded translation from this combination of transformations in order to avoid cropping issues due to the distorted digit exceeding the boundaries of the image.
We selected these pose parameter ranges in order to evaluate performance on a more challenging set of transformations than those evaluated in \citet{jaderberg2015spatial}, using a smaller training set in line with the popular Rotated MNIST dataset~\citep{larochelle2007empirical}.

For each base image, we independently sampled 8 sets of pose parameters, thus yielding 8 transformed versions of each base image.
We created 4 training sets with 10,000, 20,000, 40,000 and 80,000 examples respectively, each containing 1, 2, 4, or 8 versions of each base image.

In addition to the training sets, we generated a validation set of size 5000 using examples from the MNIST training set that were not used in the Projective MNIST training set.
Each of the images in the validation set was transformed using an independently sampled set of pose parameters sampled from the same range as the training set.

We generated a test set using all 10,000 images in the MNIST test set.
For each base image, we sampled 8 sets of pose parameters, thus yielding a test set of size 80,000 that contains 8 transformed versions of each base test 

\minihead{Preprocessing} We preprocessed the data by subtracting the mean pixel value and dividing by the standard deviation.
The mean and standard deviation were computed as scalar values, ignoring pixel locations.

\minihead{Network Architectures}
We used a baseline CNN architecture similar to the Z2CNN used in the experimental evaluation of \citet{cohen2016group}.
In all our experiments, each convolutional layer is followed by a spatial batch normalization layer.
This network has the following architecture, with output shapes listed in ($\text{channel} \times \text{height} \times \text{width}$ format):

\begin{center}
\begin{tabular}{lc}
\toprule
\textbf{Layer} & \textbf{Output Shape}\\
\midrule
input & $1\times64\times64$ \\
conv1 & $32\times62\times62$ \\
avgpool & $32\times31\times31$  \\
conv2 & $32\times29\times29$  \\
avgpool &  $32\times14\times14$ \\
conv3 & $32\times12\times12$   \\
dropout &  $32\times12\times12$ \\
conv4 & $32\times10\times10$ \\
conv5 & $32\times8\times8$ \\
conv6 & $32\times6\times6$ \\
dropout & $32\times6\times6$ \\
conv7 & $10\times4\times4$  \\
maxpool & $10$ \\
\bottomrule
\end{tabular}
\end{center}

For self-consistent pose prediction within ET layers, we used the following architecture:

\begin{center}
\begin{tabular}{lc}
\toprule
\textbf{Layer} & \textbf{Output Shape}  \\
\midrule
input & $1\times64\times64$ \\
canon. coords. & $1\times64\times64$ \\
conv1 & $32\times32\times32$ \\
conv2 & $32\times32\times32$ \\
maxpool & $32\times32$ \\
conv3 & $1\times 32$ \\
softmax & $1\times 32$ \\
centroid & $1$ \\
\bottomrule
\end{tabular}
\end{center}

In this network, the max-pooling layer eliminates the extraneous dimension in the feature map.
For example, when predicting rotation angle using polar coordinates, this operator pools over the radial dimension; the remaining feature map is then indexed by the angular coordinate.
We then pass this feature map through a softmax operator to obtain a spatial distribution, and finally compute the centroid of this distribution to obtain the predicted pose.
When a pair of pose predictions are needed (\emph{e.g.}, when predicting rotation and dilation parameters simultaneously), we use two output branches that each pools over a different dimension.

For baseline, non-equivariant pose prediction within ST layers, we used the following architecture:

\begin{center}
\begin{tabular}{lc}
\toprule
\textbf{Layer} & \textbf{Output Shape} \\
\midrule
input & $1\times64\times64$ \\
conv1 & $32\times32\times32$ \\
conv2 & $32\times32\times32$ \\
maxpool & $32\times3\times3$ \\
fc & $1$ \\
\bottomrule
\end{tabular}
\end{center}

Unlike the self-consistent pose predictor, this network does not represent the input using a canonical coordinate system.
The pose is predicted using a fully-connected layer.

\minihead{Hyperparameters}
We tuned the set of ET layers, their order, the dropout probability, the initial learning rate and the learning rate decay factor on the validation set.
ET layers were selected from subgroups of the projective group: rotation, dilation, hyperbolic rotation, $x$-shear, $x$-perspective and $y$-perspective.
The dropout probability was selected from the set $\{0.25, 0.30, 0.35, 0.40\}$.
We computed validation accuracy after each epoch of training and computed test accuracy using the network that achieved the best validation accuracy.
For optimization, we used the AMSGrad algorithm~\citep{reddi2018convergence} with a minibatch size of 128.
For the baseline and ET networks, we used an initial learning rate of $2\times10^{-3}$.
For the ST networks, we used an initial learning rate of $2\times10^{-4}$: higher learning rates resulted in unstable training and hence reduced accuracy.
We multiplicatively decayed the learning rate by $1\%$ after each epoch.
We trained all our networks for 300 epochs.

\subsection{Street View House Numbers}

\minihead{Preprocessing} For each channel, we preprocessed the data by subtracting the mean pixel value and dividing by the standard deviation.
The mean and standard deviation were computed as scalar values, ignoring pixel locations.

\minihead{Network Architectures} We used standard ResNet architectures as baseline CNNs~\citep{he2016deep}.
For pose prediction, we used the same CNN architectures as in the Projective MNIST task, but with 3 input channels.

\minihead{Hyperparameters}
As with the Projective MNIST experiments, we tuned the set of ET layers, their order, the dropout probability, the initial learning rate and the learning rate decay factor on the validation set (a 5000-example subset of the SVHN training set).
We tuned the transformation and dropout hyperparameters over the same set of possible values as for Projective MNIST.
Again, we used the AMSGrad algorithm for optimization with a minibatch size of 128.
Due to the large size ($\approx$600k examples) of the training set with the addition of the extra training images, we only trained our networks for 150 epochs in this setting.
For the remaining runs, we trained for 300 epochs.

\begin{table*}
\caption{Canonical coordinate systems implemented for our experiments with their corresponding transformation groups.}
\label{tab:coords}
\begin{center}
\begin{tabular}{lccc}
\toprule
Transformation & $T_\theta\mathbf{x}$ & $\cc_1(\mathbf{x})$ & $\cc_2(\mathbf{x})$ \\
\midrule
$x$-translation & $(x_1 + \theta, x_2)$ & \multirow{2}{*}{ $x_1$} & \multirow{2}{*}{$x_2$} \\
$y$-translation & $(x_1, x_2 + \theta)$ & \\ \midrule
Rotation & $(x_1 \cos\theta - x_2\sin\theta, x_1\sin\theta + x_2\cos\theta)$ & \multirow{2}{*}{$\tan^{-1}(x_2 / x_1)$} & \multirow{2}{*}{$\log\sqrt{x_1^2 + x_2^2}$} \\
Dilation & $(x_1 e^\theta, x_2 e^\theta)$ & \\ \midrule
$x$-scale & $(x_1 e^\theta, x_2)$ & \multirow{2}{*}{$\log x_1$} & \multirow{2}{*}{$\log x_2$} \\ 
$y$-scale & $(x_1, x_2 e^\theta)$ & & \\ \midrule
Hyperbolic Rotation & $(x_1 e^\theta, x_2 e^{-\theta})$ & $\log\sqrt{x_1 / x_2}$ & $\sqrt{x_1x_2}$ \\ \midrule
$x$-shear & $(x_1 - \theta x_2, x_2)$ & $-x_1 / x_2$ & $x_2$  \\ \midrule
$y$-shear & $(x_1, x_2 + \theta x_1)$ & $x_2 / x_1$ & $x_1$ \\ \midrule
$x$-perspective & $(x_1 / (\theta x_1 + 1), x_2 / (\theta x_1 + 1))$ & $1 / x_1$ & $\tan^{-1}(x_2 / x_1)$ \\ \midrule
$y$-perspective & $(x_1 / (\theta x_2 + 1), x_2 / (\theta x_2 + 1))$ & $1 / x_2$ & $\tan^{-1}(x_1 / x_2)$ \\
\bottomrule
\end{tabular}
\end{center}
\end{table*}

\begin{table*}[t]
\caption{Ranges of sampled transformations for Projective MNIST.}
\label{tab:ranges}
\begin{center}
\begin{tabular}{lcc}
\toprule
Transformation & $T_\theta \mathbf{x}$ & Range \\
\midrule
Rotation & $(x_1 \cos\theta - x_2\sin\theta, x_1\sin\theta + x_2\cos\theta)$ & $[-\pi, \pi]$ \\
Dilation & $(x_1 e^\theta, x_2 e^\theta)$ &  $[0, \log 2]$ \\
Hyperbolic Rotation & $(x_1 e^\theta, x_2 e^{-\theta})$ & $[-\log 1.5, \log 1.5]$ \\
$x$-shear & $(x_1 - \theta x_2, x_2)$ & $[-1.5, 1.5]$ \\
$x$-perspective & $(x_1 / (\theta_x x_1 + 1), x_2 / (\theta_x x_1 + 1))$ &\multirow{2}{*}{$\{ (\theta_x, \theta_y) \;:\; |\theta_x| + |\theta_y| \leq 0.8 \}$} \\
$y$-perspective & $(x_1 / (\theta_y x_2 + 1), x_2 / (\theta_y x_2 + 1))$ & \\
\bottomrule
\end{tabular}
\end{center}
\end{table*}

\section{Additional Experiments}

\subsection{Robustness to Unseen Transformations}

\begin{table*}
\caption{Test accuracies on affNIST. All models except (*) were trained only on randomly-translated MNIST training data and tested on affNIST images, which are MNIST test images distorted by random affine transformations.
The classification layer for (*) was trained on affNIST data using a feature extractor that was trained on MNIST.}
\label{tab:affnist}
\begin{center}
\begin{tabular}{ll}
\toprule
Method & affNIST test accuracy (\%)\\
\midrule
Standard CNN \citep{sabour2017dynamic} & 66 \\
Standard CNN (ours) & 88.3 \\
Capsule Network \citep{sabour2017dynamic} & 79 \\
Sparse Unsupervised Capsule features + SVM \citep{rawlinson2018sparse} & 90.1* \\
\midrule
Log-polar & 76.6 \\
ET-LP (translation) & 98.1 \\
ET-LP (translation + $x$-shear) & 98.3 \\
ET-Cartesian (translation + rotation/scale) & 98.2 \\
\bottomrule
\end{tabular}
\end{center}
\end{table*}

In this experiment, we evaluate the robustness of CNNs with ET layers to transformations not seen at training time.
Following the procedure of \citet{sabour2017dynamic}, we train on a variant of the MNIST training set where each digit is randomly placed on a $40\times40$ black background.
This network is then tested on the affNIST test set, a variant of the MNIST test set where each digit is transformed with a small affine transformation.\footnote{The affNIST dataset can be found at \url{https://www.cs.toronto.edu/~tijmen/affNIST/}. The transformation parameters are sampled uniformly within the following ranges: rotation in $[-20, 20]$ degrees, shear in $[-0.2, 0.2]$, vertical and horizontal scaling in $[0.8, 1.2]$. The transformed digit is placed uniformly at random on a $40\times40$ black background, subject to the constraint that no nonzero part of the digit image is cropped.}
We perform model selection against a validation set of 5000 held-out MNIST training images, each randomly placed on the $40\times40$ background but subject to no further transformations.

Our CNN baseline uses three convolutional layers with $256$, $256$ and $128$ channels, each with $5\times5$ kernels and a stride of 1. 
Each convolutional layer is followed by a batch normalization layer and a ReLU nonlinearity. 
The output of the final convolutional layer is average pooled to obtain a $128$-dimensional embedding which is mapped to $10$ classes by a fully-connected layer. We use dropout before the final classification layer.

We evaluate this CNN architecture with three ET configurations: (1) $x$- and $y$-translation followed by a transformation to log-polar coordinates,\footnote{This first ET configuration is equivalent to the Polar Transformer architecture introduced by \citet{esteves2018polar}.} (2) $x$- and $y$-translation, followed by $x$-shear, followed by a transformation to log-polar coordinates, and finally (3) $x$- and $y$-translation followed by rotation/scale, without a further log-polar transformation.

Table~\ref{tab:affnist} summarizes our results.
We find that our baseline CNN already outperforms the Capsule Network architecture from \citet{sabour2017dynamic}, while the baseline CNN over log-polar coordinates (without any ET layers) performs poorly due to the loss of translation equivariance.
The ET-augmented CNNs improve on these results, with both networks demonstrating comparable robustness to affine transformations not seen at training time.
The higher test accuracies achieved by the ET networks relative to the Capsule Network baselines reflect the stronger priors that have been built into the ET architecture---in contrast to the ET networks, the Capsule Network baselines have to learn invariances from the training data.
\end{appendix}


\end{document}